\documentclass[pdflatex,sn-mathphys-num]{sn-jnl}

\usepackage{graphicx}%
\usepackage{multirow}%
\usepackage{amsmath,amssymb,amsfonts}%
\usepackage{amsthm}%
\usepackage{mathrsfs}%
\usepackage[title]{appendix}%
\usepackage{xcolor}%
\usepackage{textcomp}%
\usepackage{manyfoot}%
\usepackage{booktabs}%
\usepackage{algorithm}%
\usepackage{algorithmicx}%
\usepackage{algpseudocode}%
\usepackage{listings}%
\usepackage{listings}%
\usepackage{lmodern} 

\begin{document}

\title{Automatic quantification of breast cancer biomarkers from multiple $^1$$^8$F-FDG PET image segmentation}

\author[1]{\fnm{Tewele W} \sur{Tareke}}\email{Tewele-Weletnsea.Tareke@u-bourgogne.fr}

\author[1,2]{\fnm{Neree} \sur{Payan}}\email{Nerea.Payan@u-bourgogne.fr}

\author[1,2]{\fnm{Alexandre} \sur{Cochet}}\email{Alexandre.Cochet@u-bourgogne.fr}

\author[3]{\fnm{Laurent} \sur{Arnould}}\email{larnould@cgfl.fr}

\author[1]{\fnm{Benoît} \sur{Presles}}\email{benoit.presles@u-bourgogne.fr}

\author[1,2]{\fnm{Jean-Marc } \sur{Vrigneaud}}\email{jmvrigneaud@cgfl.fr}

\author[1]{\fnm{Fabrice } \sur{Meriaudeau}}\email{Fabrice.Meriaudeau@u-bourgogne.fr}

\author[1,4]{\fnm{Alain} \sur{Lalande}}\email{Alain.Lalande@u-bourgogne.fr}

\affil*[1]{\orgdiv{ICMUB laboratory, UMR CNRS 6302}, \orgname{Université de Bourgogne Europe}, \city{Dijon}, \postcode{21000}, \country{France}}

\affil[2]{\orgdiv{Nuclear Medicine Department}, \orgname{Centre Georges-François Leclerc}, \city{Dijon}, \postcode{21000}, \country{France}}

\affil[3]{\orgdiv{Department of Biology and Pathology of the Tumors}, \orgname{Centre Georges-François Leclerc},  \city{Dijon}, \postcode{21000}, \country{France}}

\affil[4]{\orgdiv{Department of Medical Imaging}, \orgname{University Hospital of Dijon}, \city{Dijon}, \postcode{21000}, \country{France}}

\abstract{Background \\ Neoadjuvant chemotherapy (NAC) has become a standard clinical practice for tumor downsizing in breast cancer with $^{18}$F-FDG Positron Emission Tomography (PET) being an essential tool in treatment monitoring. Our work aims to leverage  PET imaging for the segmentation of breast lesions and the analysis of tumor metabolic response. The focus is on developing an automated system that accurately segments primary tumor regions and extracts key biomarkers from these areas to provide insights into the evolution of breast cancer following the first course of NAC.

Methods\\ 243 baseline $^{18}$F-FDG PET scans (PET$_{Bl}$) and 180 follow-up $^{18}$F-FDG PET scans (PET$_{Fu}$) were acquired before and after the first course of NAC, respectively. A deep learning-based breast tumor segmentation method was developed for the extraction of biomarkers.  Several deep learning approaches were considered.
The optimal baseline model (model trained on baseline exams) was fine-tuned on 15 follow-up exams and adapted using active learning to segment tumor areas in PET$_{Fu}$. The pipeline computes biomarkers such as maximum standardized uptake value (SUV\(_{max}\)), metabolic tumor volume (MTV), and total lesion glycolysis (TLG) to evaluate tumor evolution between PET$_{Fu}$ and PET$_{Bl}$ scans. Quality control measures were employed to identify and exclude aberrant outliers. 

Results\\ The nnUNet deep learning model outperformed in tumor segmentation on PET$_{Bl}$, achieved a Dice similarity coefficient (DSC) of \(0.89 \pm 0.04\) and a Hausdorff distance (HD) of \(3.52 \pm 0.76\) mm. After fine-tuning, the model demonstrated a DSC of \(0.78 \pm 0.03\) and a HD of \(4.95 \pm 0.12\) mm on PET$_{Fu}$ exams. Biomarkers analysis revealed very strong correlations whatever the biomarker between manually segmented and automatically predicted regions.
From these latter, the average \(\Delta SUV_{max}\), \(\Delta MTV\) and \(\Delta TLG\) between the 180 PET$_{Bl}$ and PET$_{Fu}$ scans were \(-5.22 \pm 1.55\) (p=0.001), \(-11.79 \pm 1.88\) cm\(^3\) (p=0.003) and \(-19.23 \pm 10.21 \) cm\(^3\) 
 (p=0.005), respectively. 
 
Conclusion\\ The presented approach demonstrates an automated system for breast tumor segmentation from $^{18}$F-FDG PET. Thanks to the extracted biomarkers from the segmentations on both baseline and follow-up exams, our method enables the automatic assessment of cancer progression.}

\keywords{Breast cancer,\sep $^1$$^8$F-FDG PET, \sep Biomarkers,  \sep Segmentation, \sep Deep learning, \sep Quality control, \sep Neoadjuvant chemotherapy}

\maketitle

\section{Introduction}\label{sec1}

Breast cancer is one of the most widespread and concerning forms of cancer worldwide, particularly affecting women. Lichtenstein et al. have pinpointed specific genetic variations linked to higher breast cancer risk, providing valuable insights into its causes \citep{lichtenstein2000environmental}. Neoadjuvant chemotherapy (NAC) was initially introduced as a treatment strategy for managing inflammatory or locally advanced breast cancers that are deemed inoperable at diagnosis \citep{humbert2012changes}. This approach allows for the reduction of tumor size or stage before surgical intervention and improving overall treatment outcomes. Then NAC is the first-line treatment for patients with inoperable or locally advanced breast cancer \citep{gradishar2020breast}. Minckwitz et al. \citep{von2012definition} have made the link between pathological complete response (pCR) and prognosis following NAC across various intrinsic breast cancer subtypes.

Positron Emission Tomography (PET) imaging is commonly used for assessing disease spread during diagnosis and treatment by examining the functional processes of cancerous cells within tissues. PET imaging is a crucial asset in the NAC framework, enabling early and precise assessment of treatment response in breast cancer patients \citep{paydary2019evolving,humbert2015role,han2020prognostic}. Berriolo-Riedinger et al. \citep{berriolo200718} assessed the predictive value of reduced $^{18}$F-FDG uptake in breast cancer patients undergoing NAC in relation to achieving a pCR. Cochet et al. \citep{cochet201418} evaluated the impact of $^{18}$F-FDG Positron Emission Tomography/Computed Tomography (PET/CT) on clinical management and its prognostic value in the initial staging of newly diagnosed large breast cancer. Another prospective study has evaluated the relationship between tumor blood flow, glucose metabolism (assessed by $^{18}$F-FDG PET), and proliferation and endothelial pathological markers in newly diagnosed breast cancer \citep{cochet2012evaluation}.
Biomarkers such as maximum standardized uptake value (SUV\(_{max}\)), metabolic tumor volume (MTV), and total lesion glycolysis (TLG) derived from PET imaging are essential for providing insights into tumor metabolism and size \citep{groheux2015clinical, can2022prognostic, payan2020biological, humbert2014her2}. Humbert et al. analyzed the evolution of biomarkers, such as SUV$_{max}$, to evaluate their predictive value in breast cancer treatment and their impact on clinical decision-making \citep{humbert2015role}. Similarly, a prospective study by Schwarz-Dose et al. found that biomarker measurements were significant in predicting pathological response after the first treatment cycle \citep{schwarz2009monitoring}. PET$_{Fu}$ is a widely evaluated exam for distinguishing metabolic responders from non-responders in interim or post-treatment PET scans \citep{han2020prognostic}. This review evaluated treatment response by calculating the percentage reduction in SUV$_{max}$ observed on follow-up PET scans after NAC, compared to baseline scans. This analysis was assessed both in SUV$_{max}$ changes and the pathological complete response. Gallivanone et al. \citep{gallivanone2017biomarkers} have demonstrated that biomarkers such as SUV$_{max}$, TLG, and textural features extracted from pre-treatment $^{18}$F-FDG PET and magnetic resonance imaging (MRI) are able to define patient prognosis and predict the response to NAC in breast cancer. 

Accurate breast lesion segmentation on PET scans plays a vital role in improving the precision of biomarker analysis. By separating the tumor from surrounding tissues, segmentation enhances the reliability of predictive evaluations. This process supports personalized treatment planning and the monitoring of patient progress. However, segmenting functional volumes in PET images presents a significant challenge due to several factors.
Manual delineation, while an option, is subjective, laborious, and time-consuming in medical imaging. This is particularly true for PET imaging, as well as for modalities in 3D like Computed Tomography (CT) and MRI \citep{hatt2017characterization}. 
Before 2007, the majority of breast cancer segmentation techniques consisted of using a binary threshold from PET image intensities. This thresholding often relied on methods like selecting a percentage of the SUV$_{max}$ value, setting an absolute threshold for maximum concentration of the tracer within a region of interest, or employing adaptive thresholding techniques \citep{dewalle2010new,nestle2005comparison,schaefer2008contrast,brambilla2008threshold,yaremko2005thresholding,payan2020biological}.
In the work of Payan et al. \citep{payan2020biological}, a contrast-based algorithm was employed to segment breast cancer tumors in baseline images acquired prior to the first course of NAC. However, this contrast-dependent approach has been proven to be inadequate for images with low contrast, making it unsuitable for accurately localizing tumors in images after NAC. 
Advanced image processing and computer-aided methods have emerged for processing breast cancer images \citep{liew2021review,berthon2016atlaas,li2008novel,kim2006segmentation,orlhac2014tumor}. However, their effectiveness may still be limited by factors such as tumor heterogeneity and low signal intensity, such as those affected by NAC \citep{foster2014review}.

Our study focuses on 3D PET images obtained through a specific protocol for women undergoing treatment for breast cancer 
\citep{payan2020biological}. In summary, a total of 243 PET scans at baseline and 180 scans at follow-up were included in our study. Baseline PET/CT (PET$_{Bl}$) scans were captured before the initiation of NAC or during the early stages of cancer, prior to any treatment. Follow-up PET/CT (PET$_{Fu}$) scans were acquired after the completion of the first course of NAC.
The aim of our study was to develop an automated algorithm for segmenting breast cancer lesions on PET scans. The model was trained and validated using PET$_{Bl}$ scans and applied also on PET$_{Fu}$ scans. The objective was to develop a model capable of accurately segmenting PET$_{Fu}$ scans using trained architecture from PET$_{Bl}$ scans, while also extracting imaging biomarkers. To achieve this,  UNet-based convolutional neural network architectures were employed \citep{ronneberger2015u, qamar2020variant, isensee2021nnu} specifically designed for image segmentation tasks. For PET$_{Fu}$ scans, the model developed on PET$_{Bl}$ scans was fine-tuned using a sample of PET$_{Fu}$ scans. Quality control tools were integrated into the system to ensure robustness and reliability. Moreover, our automatic segmentation method was compared with manual drawing considering inter-observer variability. Finally, three key imaging biomarkers (SUV$_{max}$, MTV, and TLG) were extracted to evaluate the model's ability to capture cancer evolution between PET$_{Bl}$ and PET$_{Fu}$ examinations.

\section{Material and Methods}
\label{sec:method}
\label{sec:dataset}
\subsection{Data acquisition}

\begin{figure*}[htbp!]
  \includegraphics[width=\textwidth]{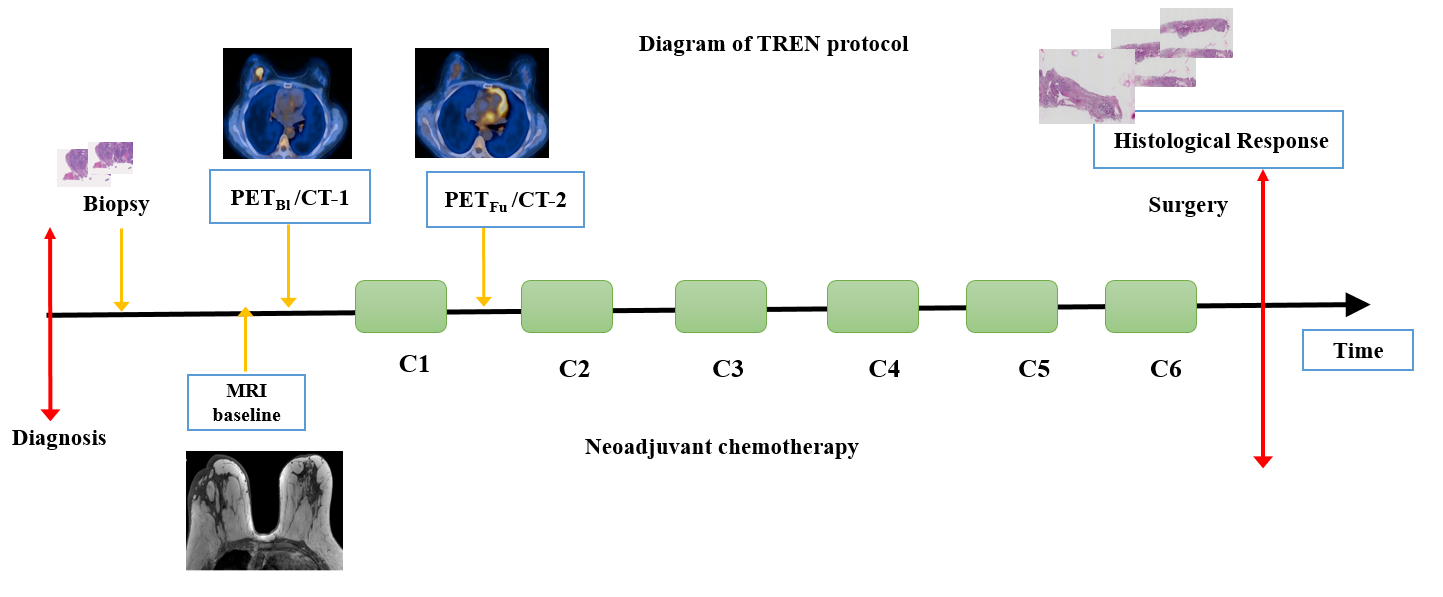}
    \caption{Diagram of our protocol. The data from the  PET$_{Bl}$ scan performed before the start of treatment are used to quantify the initial metabolism of the tumor. The data from the  PET$_{Fu}$ scan performed after the first course of chemotherapy are used to assess residual metabolism of the tumor and the metabolic response compared to the results of  PET$_{Bl}$. Magnetic Resonance Imaging (MRI) exam, which is an addition to the main protocol, was performed at baseline step.} 
    \label{fig:Tren}
\end{figure*}

\begin{figure}[htbp!]
 \centering
   \includegraphics[width= 9cm]{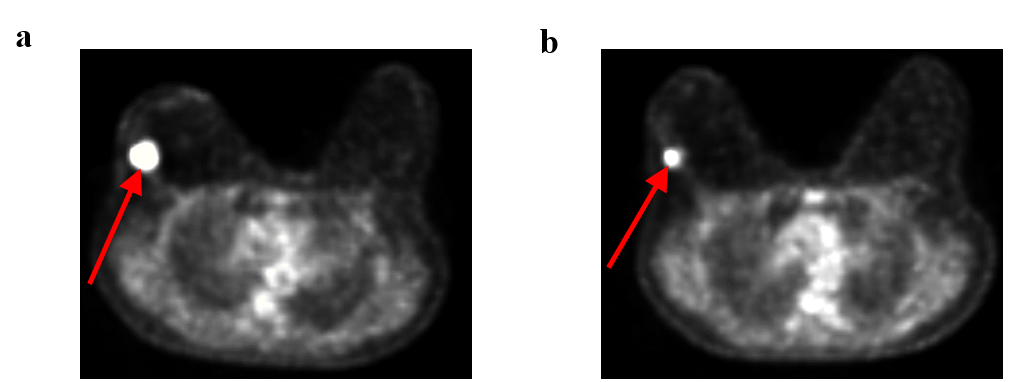}
    \caption{Axial PET images of breast tumor \textbf{a} before and \textbf{b} after first cycle of NAC for the same patient. The red arrow indicates the tumor lesion.}
    \label{fig:raw}
\end{figure}

 The data were collected from 254 patients enrolled in a clinical trial called TREN protocol (Recorded on ClinicalTrials.gov identifier: NCT02386709) \citep{payan2020biological}. This protocol aims to evaluate early metabolic and perfusion changes in invasive breast cancer patients undergoing NAC, with the objective of identifying predictive markers of tumor response and prognostic factors for recurrence risk \citep{humbert2014her2}. Successful identification of these factors allows for early adaptation of chemotherapy regimens based on marker evaluation. An overview of the protocol is illustrated in Figure \ref{fig:Tren}.
 This protocol involves administering treatment before the surgical resection, aiming to shrink tumors. Inclusion criteria involve women aged 18 or older with newly diagnosed breast cancer and a tumor diameter larger than 2 cm, eligible for neoadjuvant treatment (stage II or III). Exclusion criteria cover inflammatory tumors, distant metastasis, contraindications to treatment or surgery, pregnancy, high blood glucose level ($>$9 mmol/l),
 psychiatric illness impairing study comprehension, and unwillingness for repeated imaging.
 The protocol is a single-center, prospective, observational, non-randomized study offered to eligible patients who consent to participation \citep{payan2020biological}. The treatment protocol consists of several courses of NAC (Figure \ref{fig:Tren}). The PET$_{Bl}$ scans refer to scans acquired before the start of NAC and the PET$_{FU}$ scans are obtained after the first course of NAC. Once the inclusion and exclusion criteria were applied to the registered patients, 243 patients were retained for our study at the baseline level (Table \ref{t:dataset1}). After the first course of NAC, the number of patients decreases (missed appointment or issues during the PET scan planning). A total of 243 PET$_{Bl}$, and 180 PET$_{Fu}$ scans were therefore  considered for our study. All scans were acquired using a Gemini TruFlight PET/CT  scanner (Philips Medical Systems, Eindhoven, The Netherlands), with an axial field of view of 18 cm and a transaxial slice thickness of 4 mm. An automatic PET infusion system (Bayer Medical Care Inc., Indianola, PA, USA) was used to inject a bolus of 3MBq/kg of $^1$$^8$F-Fluorodeoxyglucose (FDG). The images were acquired in a prone position with a two-step PET/CT scan restricted on the chest. All the PET images were reconstructed using a first 3D-ordered subset iterative (OSEM) time-of-flight reconstruction technique (three iterations and 33 subsets), with image matrix sizes of 144×144 and 4 mm isotropic voxels. Emission data were corrected for random coincidences, decay, dead time, scattering and attenuation. Figure \ref{fig:raw} shows an example of PET images before and after a first course of NAC from a scan acquired in a prone position and centered on the breast. Among the 243 baseline scans, 231 tumors had already been delineated using a semi-automatic, contrast-based segmentation method \citep{schaefer2008contrast}. These delineation served as the ground truths in this study for the training of the deep learning models. Moreover, a total of 39 scans—12 at baseline and 27 at follow-up—were manually annotated by experts to serve as the ground truth for testing and fine-tuning the model (Table \ref{t:dataset1}). 

\begingroup
\setlength{\tabcolsep}{5pt} 
\renewcommand{\arraystretch}{1.9} 
\begin{table}[ht]
\centering
\small 
\caption{The distribution of the dataset. PET$_{Bl}$; Baseline exam before NAC, PET$_{FU}$; Follow-up exam after the first course of NAC, Semi-GT; Semi-annotated ground truth, Ex-GT; Expert-annotated ground truth.}
\begin{tabular}{l c c c c c c}
\hline
\textbf{Dataset} & \textbf{Total number} & \textbf{Semi-GT} & \textbf{Ex-GT} & \textbf{Training-validation} & \textbf{Fine tuning} & \textbf{Testing} \\ 
\hline
PET$_{Bl}$ exams & 243   &  231 & 12  &   231  & - & 12\\ 
PET$_{FU}$ exams & 180   & -   & 27  &   -    & 15 & 12\\  
Total exams & 423 & 231 & 39 & 231 & 15 & 24\\  
\hline
\end{tabular}
\label{t:dataset1}
\end{table}
\endgroup

\begin{figure*}[htbp]
  \includegraphics[width=\textwidth]{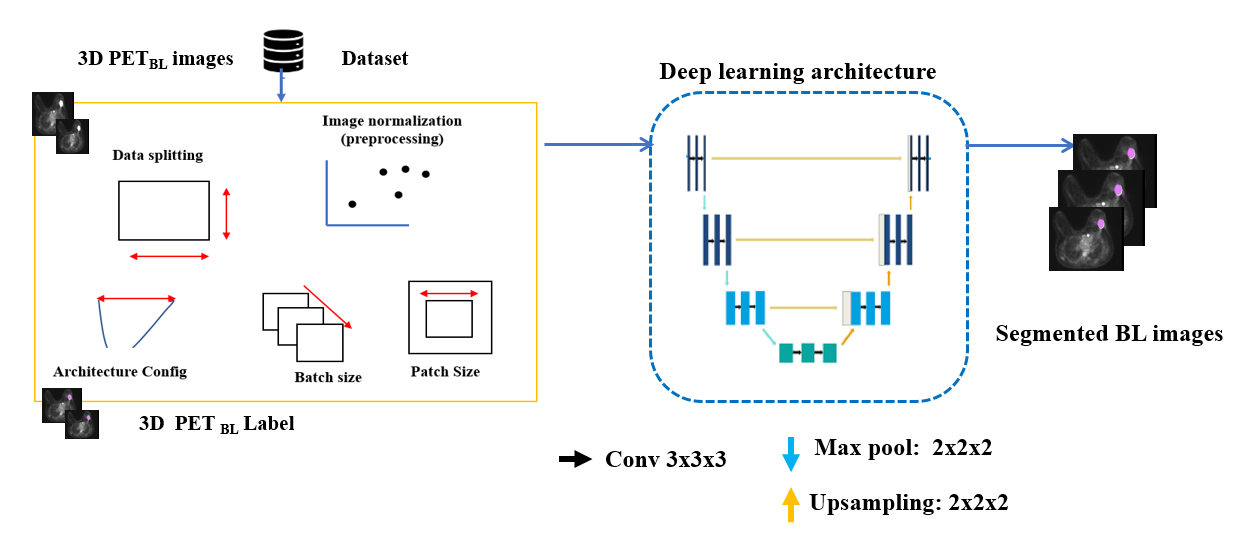}
    \caption{Pipeline of the segmentation of the tumor on PET$_{Bl}$ scans. The pre-processing part explains how the input data are organized and cleaned before being fed into the network for the segmentation of PET$_{Bl}$ scans alongside the ground truth.} 
    \label{fig:proposed}
\end{figure*}

\begin{figure*}[htbp]
  \includegraphics[width=\textwidth]{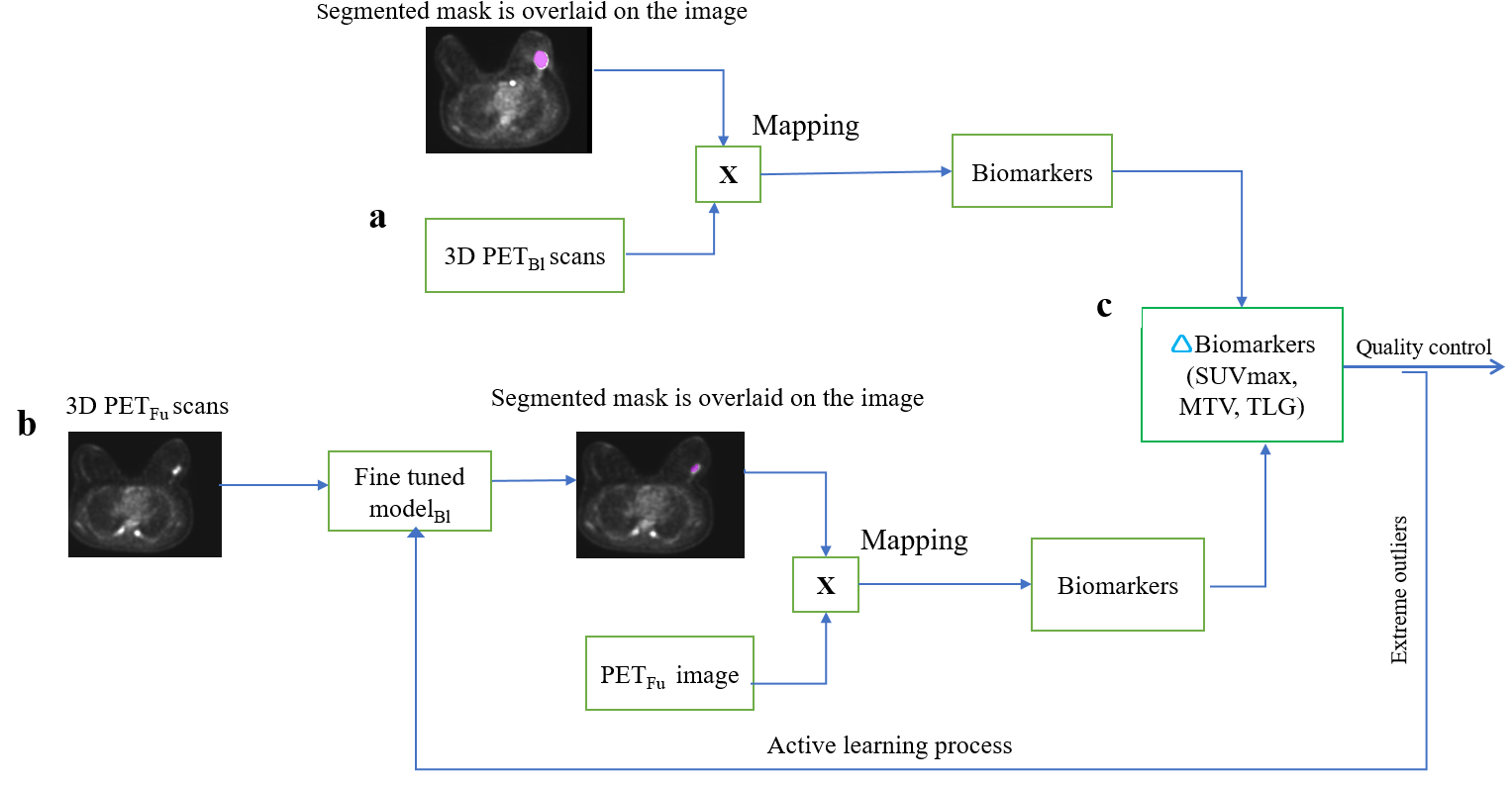}
    \caption{Pipeline for the management of the PET$_{Fu}$ scans and the biomarker extraction:
\textbf{a} Extraction of biomarkers using the segmented mask at the baseline level.
\textbf{b} Segmentation of the PET$_{Fu}$ scans using the fine-tuned baseline model (model$_{BL}$).
\textbf{c} Calculation of changes in the biomarkers between PET$_{Fu}$ and PET$_{Bl}$, such as SUV$_{max}$, to observe the impact of NAC. Active learning process is a process in which outliers identified in the PET$_{Fu}$ segmentation by the quality control system are then manually labeled to further refine the model. The terme "Mapping" corresponds to the extraction of the biomarkers from the region of interest associated to a segmented mask.} 
    \label{fig:proposed-1}
\end{figure*}

\subsection{ Deep learning models and data annotation}
Our research comprises two main phases. The first phase focuses on segmentation tasks using deep learning techniques applied to PET$_{Bl}$ scans, as depicted in Figure \ref{fig:proposed}. In the second phase, the model is fine-tuned using PET$_{Fu}$ scans to segment lesions in the PET$_{Fu}$ scans. Our approach is grounded in the robust UNet architecture, widely recognized for image segmentation in  medical imaging task \citep{ronneberger2015u, robin2021breast}.

A benchmark of classical deep learning architectures for medical image segmentation has been explored in this work, namely the 2D UNet \citep{ronneberger2015u}, 3D UNet \citep{qamar2020variant}, and nnUNet \citep{isensee2021nnu}. The nnUNet architecture enhances the traditional UNet by introducing a flexible, user-friendly framework with automated hyperparameter tuning to improve segmentation performance. 
Normalization and voxel space alignment of the volumes were implemented for all volumes as pre-processing steps to ensure that the data fed into the network is consistent and optimized for effective learning. 
Each 3D PET$_{Bl}$ scan has dimensions of 144 x 144 pixels with a depth of 66 slices. We used PET$_{Bl}$ scans to build our proposed deep learning model, considering these labels as the ground truth for training and validating the model. After selecting the best deep learning segmentation method, a fine-tunning approach was implemented. For that, 15 out of the 180 PET$_{Fu}$ scans were manually segmented by an experts and used to fine-tune the baseline model. 
The primary goal of this fine-tuning approach is to adapt the model to PET$_{Fu}$ scans by utilizing the pre-existing knowledge from the baseline model, supported by a sample of PET$_{Fu}$ scan characteristics. These 15 PET$_{Fu}$ exams were selected on the basis of a quality control process, as explained later in section \ref{subsec:QC}.
Then, to assess the model’s robustness on unseen data, the 12 reserved PET$_{Bl}$ scans and the corresponding 12 PET$_{Fu}$ scans were used to test our model and assess its generalization performances. The pipelines are illustrated in Figure \ref{fig:proposed} and Figure \ref{fig:proposed-1}.

\subsection{Loss functions}
The sum of weighted loss functions, each weighted by a certain factor \citep{marler2010weighted}, was employed. This method helped the model to adapt well to the imbalanced nature of the dataset, thereby improving the overall performance of the segmentation task. The combined weighted Focal Tversky Loss (FTL) \citep{salehi2017tversky} and Binary Cross-Entropy (BCE) loss functions \citep{ruby2020binary} was implemented, as it provides a more optimal solution compared to using BCE or FTL individually. The integration of FTL within the network effectively tackled challenges stemming from imbalanced and limited dataset.
This is achieved by assigning higher weights to minority classes, thereby prioritizing the accurate classification of rare instances and mitigating the impact of skewed class distributions. The Tversky Index (TI) is defined as:
\[
TI =  \frac{{\text{TP}}}{{\text{TP} + \alpha \cdot \text{FN} + \beta \cdot \text{FP}}}
\]
where TP is the true positive rate, FP the false positive rate, FN the false negative rate, $\alpha$ and $\beta$ control the magnitude of penalties for FPs and FNs, respectively. As TI increases, the Tversky loss function $(1 - TI)$ converges. The main difference between Tversky loss and FTL is the introduction of the focusing parameter $\gamma$ in the latter. This parameter allows the FTL to concentrate more on hard-to-classify cases (i.e. where there is hard-to-segment regions), whereas the standard Tversky loss treats all errors equally (depending on $\alpha$ and $\beta$). The focal Tversky loss is defined as:
\[
\text{FTL} = (1 - TI)^\gamma
\]
The $\gamma$ parameter controls the level of focusing on hard-to-segment regions. In another hand, Binary Cross-Entropy (BCE) is a loss function used to measure the difference between predicted probabilities and actual binary labels in binary segmentation tasks. BCE loss treats each class equally in terms of penalization, meaning that it assigns the same weight to each class during optimization, regardless of the class distribution in the dataset.  The combined loss functions is defined as:
\[
\text{Loss}_{\text{weighed}} = \epsilon \times \text{Loss}_{\text{FTL}} + (1 - \epsilon) \times \text{Loss}_{\text{BCE}}(y, \hat{y})
\]
where: \\
Loss$_{\text{FTL}}$ is the FTL function, Loss$_{\text{BCE}}$ is the BCE loss function with $y$ the true value and ($\hat{y}$) the predicted value,
\(\epsilon\) is the weight parameter representing the relative importance of the FTL compared to the BCE loss. It ranges between 0 and 1, where \(\epsilon = 0\) gives full weight to the BCE loss and \(\epsilon = 1\) gives full weight to the FTL.
\subsection{Quality control} 
\label{subsec:QC}
\begin{figure*}[htb]
  \includegraphics[width=13.5cm, height=6.0cm]{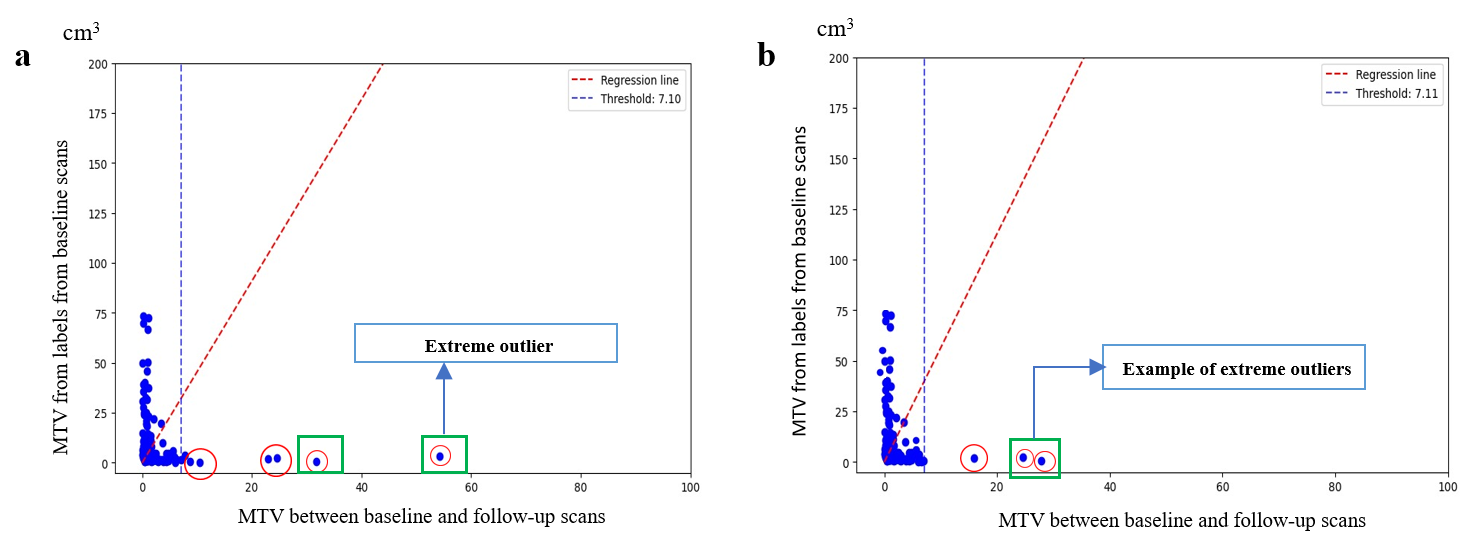}
    \caption{The distribution of MTV in PET$_{Bl}$ scans, calculated from labels, is compared to the MTV ratio between PET$_{Fu}$ and PET$_{Bl}$ cases \textbf{a} before and \textbf{b} after fine-tuning. The data points encircled by a red line are identified as outliers, while those within a rectangular green boundary are considered as extreme outliers, being the farthest from the threshold. The blue perpendicular dotted line to the x-axis represents the threshold. The regression line in \textbf{a} and \textbf{b} represents the relationship between the MTV of labeled scans and the MTV ratio between PET$_{Bl}$ and PET$_{Fu}$ scans.}
    \label{fig:threshold-tuned}
\end{figure*}

 A quality control step was integrated into the pipeline to verify segmentation accuracy and identify potential outliers. Indeed, we observed a few outliers in the changes of SUV\(_{max}\), TLG, and MTV parameters, likely due to unusually large tumor sizes. An extreme outlier is defined as a data point that lies beyond a specified threshold value within the dataset (example in Figure \ref{fig:threshold-tuned} for the MTV calculation). The purpose of this quality control approach was to first identify outliers, perform manual segmentation by experts on the corresponding scans, and finally refine the model using these new cases. This quality control step was incorporated as an active learning mechanism to enhance model performance with less labeled examples, which is especially valuable in scenarios where labeling is costly or time-intensive. The quality control process consisted of two steps. First, we assessed whether the segmented masks from PET$_{Bl}$ and PET$_{Fu}$ scans for a specific patient were localized roughly in the same area. The segmented images from PET$_{Bl}$ and PET$_{Fu}$ are binarised, with pixel values of 0 for the background and 1 for the cancer region. To objectively track whether the cancer region roughly remains in the same anatomical location over time, we divide each image of the stack of images into four quadrants.
 The center of mass of the tumor region is considered as the centroid of the region of interest.  
 If centroids from PET$_{Bl}$ and PET$_{Fu}$ fall within the same quadrant, the segmentation is considered as possible, otherwise, it is flagged as a potential error in the automatic segmentation. 
 Secondly, we calculated the ratio of follow-up to baseline MTV for each patient. An automatic, data-driven threshold was then applied to detect abnormal variations in MTV, calculated from the segmented regions before and after a first of NAC. The threshold value, determined by analyzing the relationship between the MTV extracted from baseline-labeled scans and the ratio of segmented follow-up to baseline tumor volumes across 180 scans, was set at 7.11. This threshold corresponds to the reciprocal of the mean MTV ratio. Consequently, ratio values exceeding 7.11 were identified as outliers. In practice, fifteen follow-up cases with the most extreme outliers were extracted, and manually segmented by an expert. These cases were subsequently used to fine-tune the model within the active learning pipeline. 

\subsection{Inter-observer variability}
Inter-observer variability in image segmentation refers to the differences in results obtained when different individuals (observers or annotators) segment the same images. These discrepancies can arise due to variations in the observers' interpretations, expertise, or approaches. In this study, we aimed to assess the consistency and agreement between two experienced experts, both of whom have over five years of experience in PET imaging of breast cancer. One expert (Expert-I), with more experience, was chosen as the primary observer, and then the manual annotations provided by Expert-I were considered as the reference or "ground truth" (GT) for this evaluation. Expert-II, the second observer, blindly performed manual segmentation on the same PET scans. The similarity of these two segmentation allows the assessment of the inter-observer variability. By comparing the results of the automatic segmentation with inter-observer variability, the performance of the automatic segmentation could be highlighted in relation to the expert assessments. The dataset used for this inter-observer variability study comprised 12 patients, specifically selected from the test set. 

 \subsection{Evaluation metrics}
 The evaluation of our model's performance involves the use of several metrics. The Dice similarity coefficient (DSC) is used for evaluating the similarity between the predicted and ground truth segmentation masks. It measures the overlap between the two, providing a value between 0 and 1, where a higher value indicates better agreement \citep{cappabianco2019general}. The intersection over Union (IoU), also known as the Jaccard Index, measures the overlap between the predicted segmentation and the ground truth, relative to their union \citep{cappabianco2019general}.  It ranges from 0 (no overlap) to 1 (perfect overlap). The sensitivity (True Positive Rate) is used to identify positive instances. The values of sensitivity range from 0 to 1, where 0 indicates that the model failed to identify any true positives, and 1 indicates that the model correctly identified all true positives. The hausdorff distance (HD) quantifies the maximum distance between the predicted and actual segmentation boundaries (the smaller the better) \citep{beauchemin1998hausdorff}. To evaluate inter-observer variability, we compared the overlap between lesions manually segmented by the two experts using the DSC and HD metrics.

 Three biomarkers (the SUV$_{max}$, the MTV, and the TLG) were automatically extracted from the fully-automated and the manual segmented regions at both baseline and follow-up stages from the 12 patients  used to test the models. Moreover, these metrics were automatically calculated from the automatic segmentation for the 180 cases with PET$_{Bl}$ and PET$_{Fu}$. The SUV$_{max}$ reveals the tumor's highest metabolic activity for one voxel. Meanwhile, the MTV assesses the tumor's overall metabolic volume, providing information on disease extent. The TLG combines metabolic activity and volume. The linear relationship between manually and automatically extracted biomarkers using the Pearson correlation coefficient was measured \citep{adler2010quantifying}. Paired t-test was conducted to evaluate the differences in biomarkers between patients at two different time points (baseline vs. follow-up) as our data follows a normal distribution. A p-value of less than 0.05 was considered as significant. Box plots were used  to depict the distribution of biomarkers before and after one course of NAC. The whiskers on the box plots represent the spread of data points outside the upper (third) and lower (first) quartiles.

\section{Results}
\label{sec:results}
\subsection{Hyperparameter setting}
All experiments were conducted and validated using the PET$_{Bl}$ exams mentioned in the Table \ref{t:dataset1}. The effect of varying epsilon values on the combined loss function was studied within the range [0,1].
During the hyperparameter tuning of epsilon, the values of \(\alpha\), \(\beta\), and \(\gamma\) were held constant throughout the experiments and fixed at 0.5, 0.5, and 1.0, respectively. The Table \ref{tab:alpha-values} shows that assigning a weighting factor of 0.7 to the FTL and 0.3 to the BCE loss yields optimal segmentation results.  
 We have considered different configurations of $\alpha$, $\beta$, and $\gamma$, along with $\epsilon$, to find the optimal values and improve the segmentation result. The model performed best with $\alpha = 0.7$, $\beta = 0.3$, and $\gamma = 1.5$ (Table \ref{t:comparison-0}). 

\begin{table}[ht]
\centering
\renewcommand{\arraystretch}{1.8}  
\setlength{\tabcolsep}{14pt}      
\caption{Comparison of epsilon values in the weighted sum of loss functions on the performance of the UNet-based deep learning model for the segmentation of the tumor on PET$_{Bl}$. DSC; Dice Similarity Coefficient, HD; Hausdorff Distance. Best results in bold.}
\label{tab:alpha-values}
\begin{tabular}{c c c}  
\hline
 $\epsilon$ values & DSC & HD$_{mm}$ \\
\hline
0.0 & $0.81 \pm 0.06$ & $4.81 \pm 1.03$ \\
0.2 & $0.83 \pm 0.04$ & $7.01 \pm 0.03$ \\
0.3 & $0.85 \pm 0.02$ & $3.44 \pm 0.97$ \\
0.4 & $0.79 \pm 0.09$ & $6.00 \pm 2.21$ \\
0.5 & $0.86 \pm 0.05$ & $4.33 \pm 1.23$ \\
0.6 & $0.86 \pm 0.01$ & $5.05 \pm 1.77$ \\
\textbf{0.7} & $\mathbf{ 0.89 \pm 0.04}$ & $\mathbf{3.52 \pm 0.76 }$ \\
0.9 & $0.87 \pm 0.03$ & $3.92 \pm 1.06$ \\
1.0 & $0.87 \pm 0.08$ & $4.26 \pm 1.06$ \\
\hline
\end{tabular}
\end{table}
\subsection{Segmentation results}
Among the three UNet-based deep learning architectures evaluated, the nnUNet model was chosen for its superior performance with an average DSC of $0.89 \pm 0.04$ for the best configuration using PET$_{Bl}$ dataset (Table \ref{t:comparison-0}). Figure \ref{fig:both} shows an example of lesion segmentation for two patients, each including PET$_{Bl}$ and PET$_{Fu}$ acquisition. The inter-observer variability study between experts for the manual segmentation showed a DSC score of \( 0.92 \pm 0.02 \) and HD of \( 4.02 \pm 0.21 \)(Table \ref{t:inter}). The automatic segmentation achieved  an average DSC of \( 0.89 \pm 0.04 \) and HD of \( 3.52 \pm 0.76 \) on the same PET$_{Bl}$ scans. These results indicates that the model's performance falls within the range of inter-observer variability, highlighting its consistency in replicating expert-level segmentation accuracy. Moreover, the automatic method exhibited a lower HD\(_{mm}\) of 3.52 ± 0.76, compared to 4.47 ± 0.12 for the semi-automatic approach. 
 
Table \ref{t:comparison-1} presents the baseline model's performance on PET$_{Fu}$ scans and compares the effectiveness of different adaptation strategies for segmenting breast tumors. The processes of fine-tuning and active learning led to a slight improvement in the segmentation results (average DSC of $0.78 \pm 0.03$). The output in Figure \ref{fig:quality} demonstrates the quality control step, utilizing the center of mass to assess the overlay of the volume of interest between PET$_{Bl}$ and PET$_{FU}$ scans. For the first example, the location of the ROIs and the ratio of the MTV between PET$_{FU}$ and PET$_{Bl}$ scans (\(\frac{6}{5}\)) were determined to be acceptable. For the second examples, although the ROIs were in the same quadrant, the ratio of MTV between PET$_{FU}$ and PET$_{Bl}$ scans was \(\frac{27}{1}\), which was identified as an outlier.

 \begingroup
\setlength{\tabcolsep}{6pt} 
\renewcommand{\arraystretch}{0.20} 

\begin{table*}[ht]
\small 
\centering
\caption{Comparison of baseline model performance according to the segmentation results with deep learning architectures and hyperparameter tuning techniques. Best results in bold.}
\label{t:comparison-0}
\begin{tabular}{p{0.10\linewidth} p{0.30\linewidth} p{0.12\linewidth} p{0.12\linewidth} p{0.12\linewidth} p{0.12\linewidth}}

\noalign{\smallskip} \hline \hline \noalign{\smallskip}
\vspace{1mm}\\
\textbf{Methods} & \textbf{Parameters} & \textbf{DSC} & \textbf{IOU} & \textbf{Sensitivity} & \textbf{HD}$_{mm}$ \\
\vspace{1mm}\\
\hline
\vspace{2mm}\\
2D-UNet & $\alpha=0.3, \ \beta=0.7,\ \gamma=1$ & $0.71\pm0.07$  &$0.66\pm0.02$ &$0.78\pm0.06$  & $5.60\pm0.67$  \\
\vspace{1mm}\\
3D-UNet & $\alpha=0.3, \ \beta=0.7, \ \gamma=1$ &$0.80\pm0.06$ & $0.77\pm0.03$ & $0.79\pm0.06$ &  $4.93\pm0.78$ \\
\vspace{1mm}\\
nnUNet & $\alpha=0.3, \ \beta=0.7,\ \gamma=1$ &$0.84\pm0.02$  &$0.75\pm0.06$ & $0.81\pm0.08$   &  $6.33\pm1.02$ \\
\vspace{1mm}\\
2D-UNet & $\alpha=0.5, \ \beta=0.5,\ \gamma=1$ & $0.76\pm0.03$  &$0.67\pm0.05$ &$0.77\pm0.05$  & $5.12\pm0.98$  \\
\vspace{1mm}\\
3D-UNet & $\alpha=0.5, \ \beta=0.5, \ \gamma=1$ &$0.80\pm0.07$ & $0.73\pm0.09$ & $0.81\pm0.03$ &  $6.71\pm1.05$ \\
\vspace{1mm}\\
nnUNet & $\alpha=0.5, \ \beta=0.5,\ \gamma=1$ & $0.84\pm0.05$  &$0.74\pm0.04$ & $0.83\pm0.02$   &  $6.03\pm1.30$ \\
\vspace{1mm}\\
2D-UNet & $\alpha=0.7, \ \beta=0.3, \ \gamma=1.5$ & $0.79\pm0.04$ &$0.72\pm0.01$ & $0.80\pm0.04$  & $7.01\pm0.46$  \\
\vspace{1mm}\\
3D-UNet & $\alpha=0.7, \ \beta=0.3, \ \gamma=1.5$ & $0.85\pm0.01$ & $0.77\pm0.02$ & $0.82\pm0.03$  &  $5.44\pm1.10$  \\
\vspace{1mm}\\
\textbf{nnUNet} & $\alpha=0.7, \ \beta=0.3, \ \gamma=1.5$ & $\mathbf{0.89\pm0.04}$  &  $\mathbf{0.79\pm0.07}$ & $\mathbf{0.83\pm0.05}$ &  $\mathbf{3.52\pm0.76}$ \\
\vspace{1mm}\\
\noalign{\smallskip} \hline \noalign{\smallskip}
\end{tabular}
\end{table*}

\begin{table}[ht]
\centering
\caption{Comparison of the automatic segmentation with the manual segmentation from experts as part of the inter-observer variability study (Expert-I vs. Expert-II for the manual segmentation). DSC$_{\text{avg}}$; average Dice Similarity Coefficient, HD$_{\text{mm}}$; Hausdorff Distance (in mm). GT$_{1ref}$, Ground truth from expert-I as the reference. GT$_{2}$, Ground truth from expert-II. CB, Contrast-Based}
\label{t:inter}
\scriptsize 
\renewcommand{\arraystretch}{2.5} 
\begin{tabular}{c c c c c} 
\toprule
& \multicolumn{2}{c}{\underline{Automatic Segmentation}} & \multicolumn{2}{c}{\underline{Manual GT$_{1ref}$}} \\ 
12\(_{Bl}\) cases & DSC$_{\text{avg}}$ & HD$_{\text{diff}}$ & DSC$_{\text{avg}}$ & HD$_{\text{mm}}$ \\
\midrule
Manual GT$_{1ref}$ & $0.89 \pm 0.04$ &$3.52 \pm 0.76$ & = & = \\
Manual GT$_{2}$ & $0.86 \pm 0.06$ &$6.10 \pm 1.00$ & $0.92 \pm 0.02$ &$4.02 \pm 0.21$ \\
CB semi-automatic approach & $0.91 \pm 0.02$ &$4.55 \pm 2.03$ & $0.90 \pm 0.07$ &$4.47 \pm 0.12$ \\
\bottomrule
\end{tabular}
\end{table}

\begin{figure*}[htbp!]
  \centering
  \includegraphics[width=16cm, height=10cm]{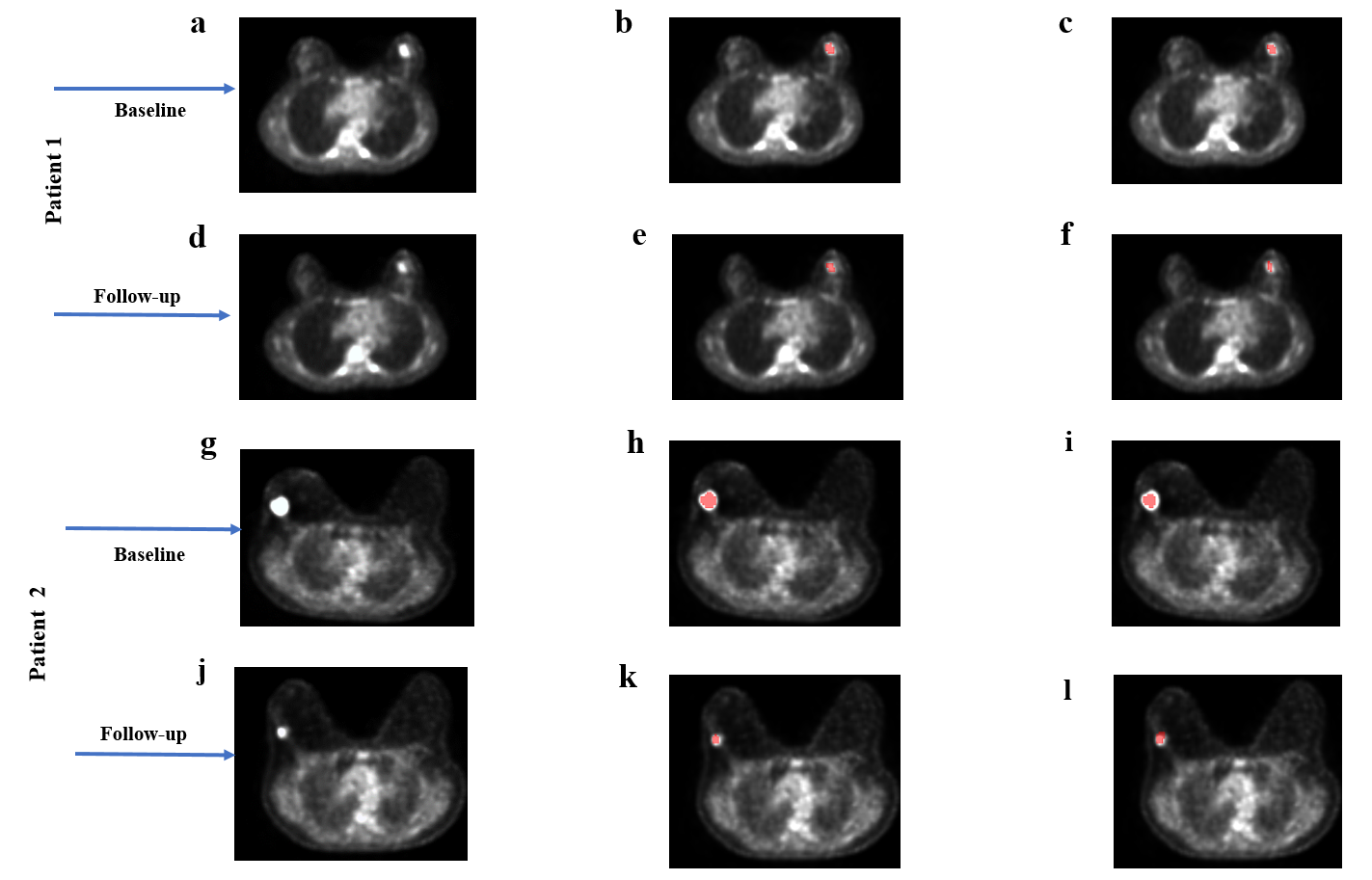}
  \caption{Segmentation results for two patients at both baseline and follow-up stages. For the patient 1, \textbf{a} and \textbf{d} represent the PET images, \textbf{b} and \textbf{e} display the corresponding ground truths in red overlaid on the images (\textbf{a}) and (\textbf{d}) respectively. \textbf{c} and \textbf{f} show the automatically segmented masks in red for images (\textbf{a}) and ( \textbf{d}) respectively. Moving on to the patient 2, \textbf{g} and \textbf{j} depict the PET images, \textbf{h} and \textbf{k} display the corresponding ground truth in red overlaid on the images (\textbf{g}) and (\textbf{j}) respectively, and \textbf{i} and \textbf{l} show the automatically segmented masks in red for images (\textbf{g}) and (\textbf{j}) respectively.}
  \label{fig:both}
\end{figure*}

A comparison of extracted biomarkers was conducted between the predicted masks from our model and the ground truth in the test dataset considering PET$_{Bl}$ and PET$_{FU}$ scans. 
 The correlation coefficient for SUV\(_{max}\) was the same for  PET$_{Bl}$ scans because there is no difference between the manual drawing and our model in SUV\(_{max}\) biomarker extraction. A mean SUV\(_{max}\) difference of 0.01 ± 0.07 was noted for PET$_{FU}$ scans. Indeed, minor discrepancies were observed due to small holes in the segmented regions.
 
  For MTV, the correlation coefficients were 0.97 and 0.93 with a mean difference of 1.57 ± 2.57 cm\(^3\) and 0.13 ± 2.97 cm\(^3\) for PET$_{Bl}$ and PET$_{FU}$ scans, respectively. Regarding TLG, the correlation coefficients were 0.91 and 0.89 with mean difference of -6.76 ± 5.65 cm\(^3\) and -3.93 ± 6.51 cm\(^3\) for PET$_{Bl}$ and for PET$_{FU}$ scans, respectively. 
 We evaluated the variation in key biomarkers between the two imaging sessions across the entire population (180 cases) using PET\(_{Bl}\) and PET\(_{FU}\) scans. To quantify changes between baseline and follow-up, we computed \(\Delta SUV_{max}\), \(\Delta MTV\), and \(\Delta TLG\). The mean difference in SUV\(_{max}\) between PET\(_{FU}\) and PET\(_{Bl}\) was -5.22 ± 1.55, corresponding a decrease of the mean of the values from 14.36 to 9.14, demonstrating statistically significant change (\(p = 0.001\)). For MTV, the average difference was -11.79 ± 1.88 cm\(^3\), corresponding to a decrease from 27.21 cm\(^3\) to 15.42 cm\(^3\), also showing statistical significance (\(p = 0.003\)). Similarly, TLG exhibited an average difference of -19.23 ± 10.21 cm\(^3\) between PET\(_{FU}\) and PET\(_{Bl}\), corresponding to a decrease from 40.69 cm\(^3\) to 21.46 cm\(^3\) with a significant \(p\)-value of \(0.005\) (Figure \ref{fig:bio-diff}).

 \begingroup
\setlength{\tabcolsep}{6pt} 
\renewcommand{\arraystretch}{1.5} 
\begin{table*}[ht]
\small 
\centering
\caption{Comparison of the performance of baseline, fine-tuned, and active learning models in segmentation tasks on PET$_{FU}$. DSC; Dice similarity coefficient, HD$_{\text{mm}}$; Hausdorff Distance (in mm), IOU; Intersection Over Union.}
\label{t:comparison-1}
\begin{tabular}{l c c c c}
\hline \hline
\textbf{Methods} & \textbf{DSC} & \textbf{IOU} & \textbf{Sensitivity} & \textbf{HD}$_{\text{mm}}$ \\ 
\hline
Baseline model & $0.75 \pm 0.07$ & $0.66 \pm 0.09$ & $0.76 \pm 0.06$ & $5.28 \pm 1.21$ \\ 
Fine-tuned model & $0.77 \pm 0.04$ & $0.66 \pm 0.04$ & $0.77 \pm 0.02$ & $5.21 \pm 0.55$ \\ 
\textbf{Active learning} & $\mathbf{0.78 \pm 0.03}$ & $\mathbf{0.69 \pm 0.05}$ & $\mathbf{0.77 \pm 0.04}$ & $\mathbf{4.95 \pm 0.12}$ \\ 
\hline
\end{tabular}
\end{table*}
\endgroup
\section{Discussion}
\label{sec:discussion}
The proposed deep learning-based method demonstrates highly promising outcomes in segmenting breast cancer lesions and extracting biomarkers from PET scans, with good performance. The model shows robustness in generalizing the method to PET$_{Fu}$ scans. Deep learning models (UNet, 3D UNet, and nnUNet) were trained and validated on 231 PET scans, and their performance was assessed using key metrics such as DSC, IOU, and HD. The nnUNet demonstrated the highest performance on the PET$_{Bl}$ scans and reasonably good performance in segmenting breast cancer lesions in PET$_{Fu}$ scans.
This difference in performance arises from the model being initially trained on the PET$_{Bl}$ dataset. Additionally, it is likely influenced by the nature of PET$_{Fu}$ scans, where a significant reduction in tumor uptake can be observed after the first course of NAC, making the segmentation task more challenging. However, our result demonstrated that refining the model through fine-tuning and active learning improved its performance in segmenting PET$_{Fu}$ scans by 3\%. This improvement is a result of refining the model using challenging cases, allowing it to better adapt to the characteristics of PET$_{Fu}$ scans as the low-uptake regions of tumor lesions. The biomarker computation also showed a high correlation between the biomarkers extracted from the predicted masks and those extracted from the ground truth masks, supporting the consistency of our method. Semi-automatic contrast-based method \citep{schaefer2008contrast} showed  comparable performance on PET$_{Bl}$ with a mean DSC of \(0.90 \pm 0.02\) whereas our method achieved a mean DSC of \(0.89 \pm 0.04\) on PET$_{Bl}$ scans. Our approach achieved similar accuracy to semi-automatic methods while significantly improving consistency and boundary precision. On PET$_{Fu}$ scans, our method maintained a DSC of \(0.78 \pm 0.03\), showing its robustness, especially in challenging cases such as low-intensity images, where the semi-automatic contrast-based method struggled to reliably segment the lesion. The proposed system seems to perform comparably to manual and semi-automatic approaches while reducing manual intervention, minimizing human error and variability, and efficiently managing large volumes of data.

Regarding these segmentation results, our method demonstrates performance comparable to previous studies on lesion segmentation using PET images \citep{moreau2021automatic, myronenko2022automated, moreau2020deep, hatt2017characterization}. Our segmentation performance is comparable to that reported by Moreau et al. \citep{moreau2021automatic}, who achieved a mean DSC of \(0.78 \pm 0.07\) for  PET$_{Bl}$ scans and \(0.56 \pm 0.11\) for  PET$_{Fu}$ scans using a UNet network to segment metastatic breast cancer lesions. Our model presents overall better results, particularly on PET$_{Fu}$ scans. However, the segmentation of metastatic lesions poses further challenges due to their variable sizes and locations, which may reduce the overall performance of the proposed method. Blanc-Durand et al. reported a mean DSC of 0.79 for the segmentation of diffuse large B-cell lymphoma lesions \citep{blanc2021fully}. Additionally, Gudi et al. emphasized the difficulty of head and neck tumor segmentation, achieving DSC of 0.69 on PET/CT scans \citep{gudi2017interobserver}. 
Vogl et al. \citep{vogl2019automatic} explored a data-driven machine learning approach for a Computer-Aided Diagnosis system using dynamic contrast-enhanced-MRI, diffusion-weighted imaging, and \(^{18}\)F-FDG PET. They employed a random forest classifier combined with multi-parametric PET/MRI intensity-based features for breast lesion segmentation, achieving a DSC of 0.67. They employed a leave-one-out cross-validation approach to evaluate lesion segmentation performance on a dataset of 34 cases.
A study conducted by Qiao et al. \citep{qiao2022improving} on breast tumor segmentation in PET imaging through attentive transformation-based normalization yielded promising results, achieving a DSC of \(0.86 \pm 0.06\). They conducted their evaluation using a subset of 10 scans for testing, selected from a total of 54 scans in their dataset.  
Building upon their findings, our research endeavors to extend the boundaries of tumor segmentation by integrating advanced deep learning and combined loss function techniques. 
\begin{figure*}[htbp!]
  \includegraphics[width=14cm, height=7.5cm]{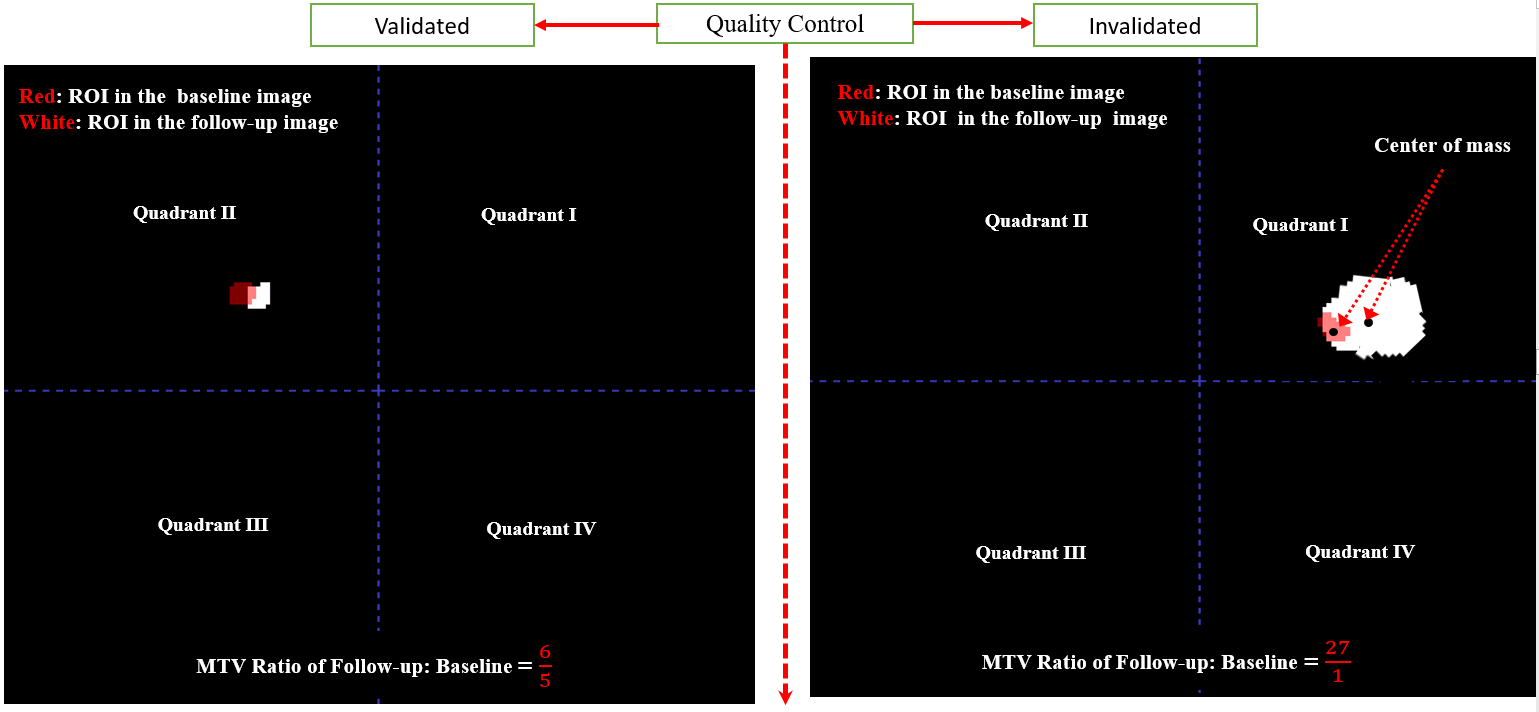}
    \caption{Example of a quality control method from predicted tumor areas from  PET$_{Bl}$ and  PET$_{Fu}$ from the same patient. In the \textbf{Validated scenario on the left}, both ROIs automatically predicted are located in the same quadrant, with the MTV ratio between them being below the specified threshold. In the \textbf{Invalidated scenario on the right}, although the ROIs are in the same quadrant, the MTV ratio exceeds the threshold. This is one of the extreme outliers identified by our quality control system.}
    \label{fig:quality}
\end{figure*}

By accurately segmenting the breast lesions at baseline and follow up stages, we were able to extract relevant biomarkers from the segmented regions to track breast cancer progression after the first course of NAC. Several research studies have investigated biomarkers to predict the response to NAC in primary breast cancer. In a study of Humbert et al. \citep{humbert2015identification}, the value of metabolic tumor response, assessed using \(^{18}\)F-FDG-PET, was evaluated for its ability to predict the pathological complete response in women with triple-negative breast cancer. The study found that combining a low metabolic response (\(\Delta\)SUV$_{max}$ \(\leq\) -50\%) with positive epidermal growth factor receptor status predicted non-pCR with 92\% of accuracy.  
Furthermore, Humbert et al. \citep{humbert2014her2} demonstrated that low tumor metabolism after the first cycle of NAC (SUV$_{max}$ $<$ 2.1) strongly predicts pCR in HER2-positive breast cancer, suggesting its potential as an early indicator of treatment success.  Woolf et al. \citep{woolf2014evaluation} reported a mean baseline SUV$_{max}$ of 7.3, which decreased to 4.62 after one cycle of NAC, reflecting a reduction of 2.68 (36.3\%). Their findings support that SUV$_{max}$ obtained from Fluorothymidine-PET-CT scans is a well-known marker for assessing cellular proliferation. In our study, we quantified the biomarkers before and after the first course of NAC. We analyzed clinical parameters for 180 cases and observed a notable reduction in metabolic, volumetric, and TLG biomarkers (Figure \ref{fig:bio-diff}). In details, metabolic activity, represented by SUVmax, decreased from 14.36 at baseline to 9.14 at follow-up, reflecting a reduction of 5.22 (36.4\%). Similarly, MTV declined from 27.21 cm³ at baseline to 15.42 cm³ at follow-up, corresponding to a decrease of 11.79 cm³ (43.3\%). TLG exhibited the most substantial reduction, dropping from 40.69 cm³ at baseline to 21.46 cm³ at follow-up, marking a  decrease of 19.23 cm³ (47.3\%). The observed variations reinforce the utility of PET-derived quantitative biomarkers as indicators of therapeutic efficacy, supporting their integration into clinical decision-making for personalized treatment strategies. Han et al. \citep{han2020prognostic} investigated the percentage change in SUV$_{max}$ (\%$\Delta$SUV$_{max}$) and highlighted its role as a widely studied parameter for distinguishing metabolic responders from non-responders in interim or post-treatment PET scans.  
Regular monitoring and integration of clinical and imaging data are crucial for comprehensive evaluation and informed decision-making.  In breast cancer patients undergoing NAC, a pCR can be accurately predicted by a reduction in \(^{18}\)F-FDG PET uptake after just one course of chemotherapy \citep{berriolo200718}. This study demonstrates that a relative decrease of 60\% in SUV\(_{max}\) or 50\% in SUV\(_{average}\) serves as a predictive threshold for achieving pCR. These findings could participate to the patient management by identifying ineffective chemotherapy early or supporting decisions to proceed with dose-intensive preoperative chemotherapy in patients who are responding. Végran et al. \citep{vegran2009gene} explored the potential of microarray analyses to identify markers, such as genes, associated with pCR. Incorporating additional markers like TLG and SUV$_{max}$ could enhance the accuracy of pCR prediction. Guiu et al. \citep{guiu2013pathological} showed also that pCR serves as a prognostic factor, particularly in patients with hormone receptor-positive tumors. 

While our study presents promising findings, it is crucial to acknowledge certain limitations. The proposed pipeline should be validated on larger and more diverse datasets, considering variations in imaging protocols and patient populations.

While our focus was on $^{18}$F-FDG PET imaging, incorporating multimodal imaging could enhance the depth of our investigation. For each PET scan, a corresponding CT scan was acquired simultaneously. Incorporating information from the CT scan could boost the performance of the proposed system. Future research could explore extracting additional information (like textures) from segmented masks, combined with clinical data, to predict treatment response and patient monitoring strategies. 

\begin{figure*}[htbp!]
\centering
  \includegraphics[width=8cm, height=11.5cm]{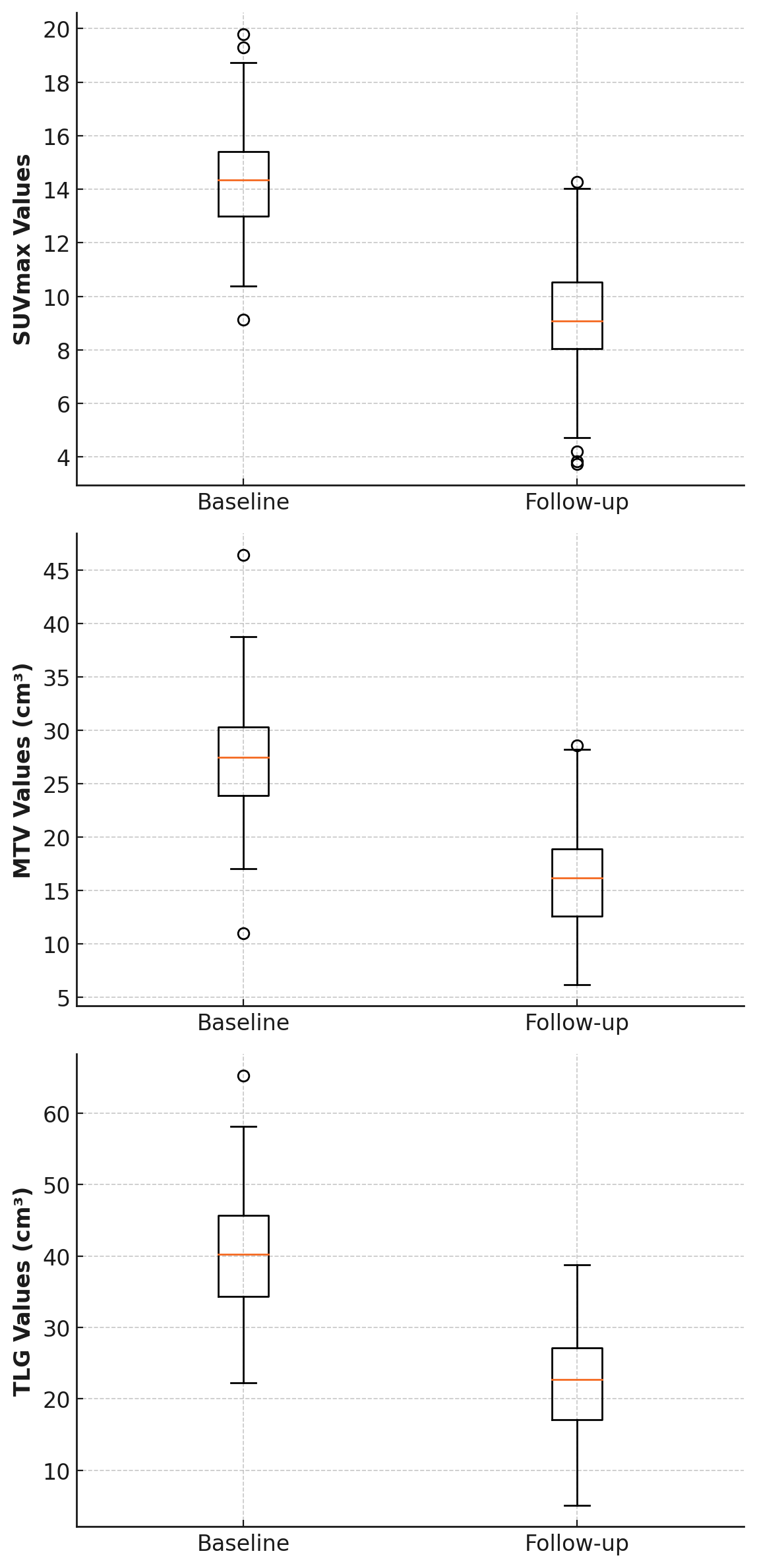}
    \caption{Box plots comparing biomarker values for SUV$_{max}$, MTV, and TLG at baseline and follow-up stages for 180 patients. Each box represents the interquartile range (IQR), with the red line inside indicating the median value. Measurements at baseline stage (left) are compared to measurements at follow-up stage (right) for each parameter, illustrating changes after NAC based on the biomarker values extracted from our system.}
    \label{fig:bio-diff}
\end{figure*}
\section{Conclusion}
\label{sec:conclusions}
In conclusion, this study presents an approach for the automatic segmentation and biomarker computation of breast cancer lesions from  $^1$$^8$F-FDG PET scans. The proposed pipeline is based on deep learning architectures and a weighted sum of loss functions. It demonstrated very good performance in accurately segmenting tumors from PET scans at both the baseline stage and follow-up level after one course of NAC. Our pipeline has successfully employed fine-tuning processes and active learning methods. Moreover, quality control measures were implemented to ensure the consistency and validity of the segmentation results.  This approach holds promise for improving the efficiency of tracking disease progression, thereby enhancing personalized treatment planning and monitoring. Beyond the calculated biomarkers, clinical data and additional parameters could be considered to build a predictive model for treatment response and survival rates. Our developed tool could be integrated into the treatment pipeline for patients with breast cancer. 

\setlength{\parindent}{0pt}
\vspace{5mm}

\textbf{Abbreviations}

BCE\hspace*{0.3cm} Binary Cross-Entropy\\
CT \hspace*{0.3cm} Computed Tomography\\
DSC\hspace*{0.3cm}  Dice Similarity Coefficient\\
FDG\hspace * {0.3 cm} Fluorodeoxyglucose\\
FN  \hspace*{0.3cm}  False Negative\\
FP  \hspace*{0.3cm}  False Positive\\
FTL\hspace*{0.3cm} Focal Tversky Loss\\
GT  \hspace*{0.3cm}  Ground Truth\\
HD \hspace*{0.3cm}Hausdorff Distance\\
IoU\hspace*{0.3cm} Intersection over Union\\
Model$_{BL}$   \hspace*{0.3cm}  Baseline Model\\
MRI \hspace*{0.3cm}  Magnetic Resonance Imaging\\
MTV \hspace*{0.3cm} Metabolic Tumor Volume\\
NAC \hspace*{0.3cm}Neoadjuvant chemotherapy\\
nnUNet \hspace *{ 0.3 cm} No-new-UNet\\
pCR  \hspace*{0.3cm} Pathological Complete Response \\
PET \hspace*{0.3cm} Positron Emission Tomography\\
PET$_{Bl}$\hspace*{0.3cm}  Positron Emission Tomography before first course of NAC (baseline)\\
PET$_{Fu}$ \hspace*{0.3cm} Positron Emission Tomography after first course of NAC (follow-up)\\
ROI   \hspace*{0.3cm} Region of Interest \\
SUV\hspace*{0.3cm} Standardized Uptake Value\\
SUV\(_{max}\) \hspace*{0.3cm} Maximum Standardized Uptake Value\\
TI\hspace*{0.3cm} Tversky Index\\
TN  \hspace*{0.3cm}  True Negative\\
TP  \hspace*{0.3cm}  True Positive\\

\textbf{Acknowledgements}
\\
None
\vspace{5mm}

\textbf{Author contributions}
 
TT: Conceived and designed the study, performed data analysis, interpreted the data, developed the methodology, and drafted the manuscript.  
NP: Contributed to the conception of the study, interpreted the data, and carefully reviewed and revised the manuscript.  
BP, AC, JMV, and LA: Responsible for patient inclusion, clinical interpretation, and carefully reviewed and revised the manuscript.  
FM and AL: Conceived and designed the study, interpreted the data, and carefully reviewed and revised the manuscript. 
\vspace{5mm}

\textbf{Funding}

This study was funded by Marie Skodowska-curie doctoral network actions (horizon-msca-2021-dn-01-01) with a grant number of 101073222.
\vspace{5mm}

\textbf{Availability of data and materials}

The datasets used and analysed during the current study are not publicly
available but are available from the corresponding author on reasonable
request.
\vspace{5mm}

\textbf{Declaration of interest}

The authors declare that there are no conflicts of interest that could be perceived as prejudicing the impartiality of the reported research.
\vspace{5mm}
 
\textbf{Ethics approval and consent to participate}
\\
This study adhered to the principles outlined in the Declaration of Helsinki and received approval from the institutional ethics committee of the Georges-François Leclerc Center. The institutional review board approved this prospective study as part of standard care (NCT02386709). Patient non-opposition was documented in source records by the medical team and served as the basis for informed consent.
\vspace{5mm}

\textbf{Consent for publication}
\\
Not applicable
\vspace{5mm}

\textbf{Competing interest}
\\
The authors declare that they have no conflicting financial interests or personal relationships that could have influenced the work presented in this paper.
\bibliography{sn-bibliography}

\end{document}